\pdfoutput=1

\documentclass[11pt]{article}

\usepackage[]{acl}

\usepackage{times}
\usepackage{latexsym}

\usepackage[T1]{fontenc}

\usepackage[utf8]{inputenc}

\usepackage{microtype}

\usepackage{inconsolata}

\usepackage{graphicx}
\usepackage{paralist}
\usepackage{url}
\usepackage{multirow}
\usepackage{booktabs}

\usepackage{amssymb}
\usepackage{array}

\title{Real-World Summarization:

When Evaluation Reaches Its Limits}

\author{Patrícia Schmidtová \\
  Charles University\thanks{Work done while on internship at trivago.}  \\
  \hspace{2.5cm}\texttt{\{schmidtova,odusek\}@ufal.mff.cuni.cz}\hspace{-2.5cm} \\\And
  Ondřej Dušek \\
  Charles University \\
   \\\And
  Saad Mahamood \\
  trivago \\
  \texttt{saad.mahamood@trivago.com} \\
  }

\begin{document}
\maketitle
\begin{abstract}
We examine evaluation of faithfulness to input data in the context of hotel highlights—brief LLM-generated summaries that capture unique features of accommodations. 
Through human evaluation campaigns involving categorical error assessment and span-level annotation, we compare traditional metrics, trainable methods, and LLM-as-a-judge approaches.
Our findings reveal that simpler metrics like word overlap correlate surprisingly well with human judgments (r=0.63), often outperforming more complex methods when applied to out-of-domain data.
We further demonstrate that while LLMs can generate high-quality highlights, they prove unreliable for evaluation as they tend to severely under- or over-annotate. Our analysis of real-world business impacts shows incorrect and non-checkable information pose the greatest risks.
We also highlight challenges in crowdsourced evaluations.

\end{abstract}

\vspace{-0.5em}
\section{Introduction}
\vspace{-0.5em}

Instruction-tuned large language models (LLMs) have become ubiquitous in natural language processing  \cite{qin2024largelanguagemodelsmeet}. They proved very capable and versatile, with simple prompting achieving various tasks, 
avoiding the need to produce costly in-domain training datasets required for most previous approaches \cite{wei_finetuned_2022}.
However, LLMs are known to have issues with faithfulness of their outputs to the input, where they produce text not grounded in the input prompt (\emph{hallucinations}; \citealp{ji_survey_2023}).

Evaluating faithfulness is therefore crucial, especially without human-written references, which are expensive and risk  leakage into LLMs' training data \cite{oren_proving_2024}.
However, traditional evaluation metrics for NLP show low correlation with human judgments \cite{novikova-etal-2017-need} and overt reliance on surface similarities \cite{gehrmann-2023-repairing}. Despite this, most works in the field still rely on them \cite{schmidtova-etal-2024-automatic-metrics}.
A new alternative is using LLMs themselves to evaluate generated outputs \cite{gu2025surveyllmasajudge,bavaresco2024llms}. While promising, LLMs may show self-bias \cite{koo-etal-2024-benchmarking} and performance may vary across domains.

We focus on faithfulness evaluation of text summarization using a case study of generating short  highlights from hotel descriptions and reviews \cite{kamath-etal-2024-generating-hotel}. 
We are concerned with the following research questions: (1) How well do faithfulness metrics generalize to unseen domains? (2) Can we evaluate faithfulness in a referenceless scenario? (3) Can LLMs be used as judges in this setting? and (4) What is the estimated business impact of these errors?

In response to these questions, we present the following contributions:
\begin{compactitem}
\item We demonstrate that simple metrics (e.g., word overlap) outperform most trainable methods on out-of-domain data. Sophisticated methods may not generalize effectively.
\item We validate multiple referenceless evaluation methods against human annotations, finding several methods that correlate well with human judgments (up to r=0.67).
\item We provide empirical evidence cautioning against uncritical use of LLM-as-a-judge approaches, revealing systematic tendencies to either over- or under-annotate errors, depending on the model used.
\end{compactitem}

\vspace{-0.5em}
\section{Task Description}
\vspace{-0.5em}
Hotel Highlights \cite{kamath-etal-2024-generating-hotel} are brief LLM-generated summaries that capture unique features of hotels based on their descriptions and reviews.
These highlights help travelers select appropriate accommodations without reading numerous reviews and lengthy descriptions. 
Consequently, ensuring highlight accuracy is essential.

The highlights intentionally present hotels in a positive light, often containing subjective phrases like `a local gem' that, while not explicitly supported by the input text, are not considered errors.
This characteristic introduces additional evaluation challenges.

The objective of the task is not to produce a single gold-standard summary, but rather to generate a diverse set of approximately 10 highlights that can be ranked and sampled.
We therefore focus primarily on verifying whether the information in each highlight is properly grounded in the source description. 
Table \ref{tab:hotel_description} demonstrates an example hotel description with two corresponding highlights.

\vspace{-0.5em}
\begin{table*}
    \centering\small
    \begin{tabular}{p{0.97\textwidth}}
        \toprule
        \textbf{Description:} \textit{Just a 5-minute walk from Mall of the Emirates, DoubleTree by Hilton Hotel and Residences Dubai offers modern accommodations. [...] The hotel is 7.0 km from Dubai Marina and 12.1 km from Dubai Mall. Dubai International Airport is 30 minutes away by car.} \\
        \midrule
        \textbf{H1:} \textit{Shop in the Mall of the Emirates thanks to the hotel's convenient location.} \\
        \midrule
        \textbf{H2:} \textit{Enjoy wonderful views across \textcolor{red}{\underline{the Hudson River to New Jersey and Liberty Island}} from select suites.} \\
        \bottomrule
    \end{tabular}
    \caption{Hotel description excerpt with two corresponding highlights. H1 is correct but H2 contains incorrect information (highlighted in red). H2 was used as one of the attention checks.}
    \label{tab:hotel_description}
\end{table*}

\section{Human Evaluation}
\label{sec:human}

\vspace{-0.5em}
\subsection{Categorical Annotations}
\label{sec:categorical_human}
We utilized a dataset from prior work \citep{kamath-etal-2024-generating-hotel}, containing 120 highlight-summary pairs.
Each pair received annotations from 30 evaluators, categorized as: \emph{no errors}, \emph{hallucination}, \emph{contradiction}, or \emph{both types} of errors.
Each annotator completed one attention check -- an example that contained a very prominent hallucination.
For the purposes of this paper, we filtered out all of the annotators who did not pass this attention check.
After filtering, 19 to 22 judgments were available for each example.
The annotators frequently do not fully agree on the presence of hallucination -- we choose to interpret this as a signal and operate with the percentage of eligible annotators who found a hallucination in the example.

\vspace{-0.5em}
\subsection{Span Annotations}
We implemented span-level error annotation to obtain more explainable and actionable feedback on highlight quality.
Through manual inspection, we identified three error types: \emph{non-checkable} (information not included in the description), \emph{misleading} (information taken out of context), and \emph{incorrect} (contradicting the description).

\vspace{-0.5em}
\paragraph{Annotators}
We gathered annotations from 124 crowd workers recruited via Prolific across 496 description-highlight pairs.
The annotators were native English speakers from the United Kingdom or United States with >90\% approval rates.

Before launching the evaluation, we conducted three pilot studies to assess guideline clarity, cognitive load, and expected completion time.
The total annotation cost, including pilots, approached £800.

\vspace{-0.5em}
\paragraph{Method}
Each annotator evaluated 8 sampled pairs plus two manually selected attention checks, totaling 10 examples.
The annotations were collected using the Factgenie interface \citep{kasner-etal-2024-factgenie-framework}
Each example was annotated by two annotators.
Based on pilot findings, we ran separate campaigns for shorter (Group A) and longer descriptions (Group B), to provide consistent time estimates for completion.
The annotators were positively motivated to focus on quality by a bonus payment to those passing our two attention checks.

\vspace{-0.5em}
\paragraph{Results}
The annotations (summarized in Figure~\ref{fig:span_error_distribution}) indicate 58\% of highlights are error-free, while the remainder contain non-checkable (20\%), misleading (19\%), or incorrect (7\%) content.\footnote{Individual highlights may contain multiple error types.}

\vspace{-0.5em}
\paragraph{Quality}
Only 24\% of annotators passed our attention checks, with over-annotation emerging as a common issue.
This was confirmed by internal domain experts who analyzed 20 annotations per error type and found slightly over half of the annotated spans contained no actual errors.
This problem was more pronounced in group B.
We suspect this stems from source text length—when annotators struggled to quickly locate information in the text, they marked the spans containing this information as errors.
This occurred despite explicit encouragement to use Ctrl+F for efficiently locating information.

\vspace{-0.5em}
\subsection{Estimation of Real-World Impacts}

On a sample of 60 error span annotations described above, we determined that 32 have \emph{no business impact} (no actual error), 13 present \emph{low business impact} (clients unlikely to complain about being misled), 13 show \emph{medium business impact} (clients might complain without requesting compensation), and only 2 indicate \emph{high business impact} (clients likely to request compensation or a significant reputation risk).
Incorrect information most frequently causes a higher business impact, followed by non-checkable information.

\vspace{-0.5em}
\section{Validating Quality Estimation Methods}
\label{sec:automatic}
\vspace{-0.5em}
\subsection{Example-level Binary Classification}
We experimented with automatically determining whether a given example contains a semantic error or not.
Using data described in Section \ref{sec:categorical_human}, we calculate the Spearman rank correlation between automatic metric scores and the percentage of annotators who believe there is an error in a given highlight.
By doing so, we are aiming to capture the subjectivity and uncertainty as a signal -- if most annotators agree there is (not) an error, then the automatic metric should agree with the majority to be considered reliable.
We consider the following metric types:

\begin{table}[t]
\centering\small
\begin{tabular}{llr}
\toprule
\textbf{Type} & \textbf{Metric} & \textbf{Corr. (Spearman)} \\
\midrule
\multirow{3}{*}{O} 
  & Form / Lemma overlap        & \textbf{0.63} / 0.62 \\
  & Noun overlap        & 0.41 \\
  & Adjective overlap   & 0.55 \\
 
\midrule
\multirow{2}{*}{N} 
  & BLEU                & 0.51 \\
  & ROUGE-L (P / R / F)   & 0.38 / \textbf{0.56} / 0.41  \\
\midrule
\multirow{3}{*}{T}
  & NLI-entailment & \textbf{0.67} \\
  & BertScore & 0.57 \\
  & LaBSE & 0.12 \\
\bottomrule
\end{tabular}
\caption{Correlation for overlap-based methods (O), n-gram overlap methods (N), and trainable methods (T). 
This table includes a small selection of the explored methods, see Table \ref{tab:hallucination_and_error_correlations} in the Appendix for full results.
}
\label{tab:corrs}
\end{table}

\vspace{-0.5em}
\paragraph{Single Word Overlap}
The simplest method we consider -- overlap of word forms between the highlights and descriptions -- proved to reach the highest correlations with the human annotation.
We tested several variations, shown in Table~\ref{tab:corrs}. 
Word overlap can easily be confused by phenomena such as negation; however, it is cheap and quick to calculate.

\vspace{-0.5em}
\paragraph{N-Gram Overlap}
We measured BLEU \citep{post-2018-call} (without brevity penalty) and ROUGE-L \citep{lin-2004-rouge} between the highlights and the descriptions.
They reach correlations with human judgments comparable to single-word methods while being more robust, because they consider a longer combination of n-grams.

\vspace{-0.5em}
\paragraph{Natural Language Inference (NLI)} was demonstrated to work as a referenceless metric for semantic accuracy in data-to-text generation \cite{dusek-kasner-2020-evaluating} and summarization \cite{maynez-etal-2020-faithfulness}.
If the generated summary is entailed by the source description, the intended meaning was likely preserved.
On the contrary, if the summary is not entailed by the description, then it is likely there is a semantic error.

We performed initial experiments with an older NLI model, DeBERTa v.3 \citep{he2021deberta}
\footnote{\url{https://huggingface.co/cross-encoder/nli-deberta-v3-base}} to measure the entailment likelihood between the source and the highlight.
The results did not seem promising -- for the vast majority of samples, the likelihood of entailment was very close to 0 or 1, with only a 0.08 correlation with human judgments on hallucinations.

However, using a newer model trained on more data, ModernBERT \citep{sileo-2024-tasksource}\footnote{\url{https://huggingface.co/tasksource/ModernBERT-base-nli}}, proved to be helpful as we reached a correlation of 0.67 with human judgment on the categorical set.

\vspace{-0.5em}
\paragraph{Text Embedding Similarity}
We have experimented with 6 different embedding models and found that BertScore \citep{bert-score} reached a moderate correlation with human judgment. By comparison, the next similarity-based measure -- cosine similarity of LaBSE embeddings \citep{feng-etal-2022-language} reached an unsatisfying correlation of 0.12.
The issue with these measures was that the similarity always stays high due to thematic closeness but is unable to reflect small pieces of unsupported information in the highlights.

\vspace{-0.5em}
\subsection{Locating the Error and its Severity}

\begin{table*}[htbp]
\centering
\small
\begin{tabular}{l>{\hspace{-2mm}}l>{\hspace{-2mm}}r>{\hspace{-1mm}}r>{\hspace{-1mm}}r>{\hspace{-1mm}}r>{\hspace{-1mm}}r>{\hspace{-1mm}}r>{\hspace{-1mm}}r>{\hspace{-1mm}}r>{\hspace{-1mm}}r}
\toprule
\textbf{Reference} & \textbf{Hypothesis} & \textbf{Ref Ct} & \textbf{Hyp Ct} & \textbf{Prec (H)} & \textbf{Rec (H)} & \textbf{F1 (H)} & \textbf{Prec (S)} & \textbf{Rec (S)} & \textbf{F1 (S)} \\
\midrule
Human A (longer)   & Human B (longer)   & 176 & 154 & 0.179 & 0.223 & \textbf{0.199} & 0.307 & 0.382 & \textbf{0.340} \\
Human A (longer)   & Gemma3             & 176 &  85 & 0.082 & 0.060 & 0.069 & 0.185 & 0.136 & 0.157 \\
Human A (longer)   & GPT-4o     & 176 & 258 & 0.068 & 0.141 & 0.092 & 0.143 & 0.293 & 0.192 \\
Human A (shorter)  & Human B (shorter)  & 144 & 161 & 0.107 & 0.104 & 0.106 & 0.288 & 0.280 & 0.284 \\
Human A (shorter)  & Gemma3             & 144 &  93 & 0.085 & 0.066 & 0.074 & 0.226 & 0.175 & 0.197 \\
Human A (shorter)  & GPT-4o    & 144 & 298 & 0.108 & \textbf{0.232} & 0.147 & 0.198 & 0.425 & 0.270 \\
Human B (longer)   & Human A (longer)   & 154 & 176 & \textbf{0.217} & 0.175 & 0.194 & \textbf{0.376} & 0.302 & 0.335 \\
Human B (longer)   & Gemma3             & 154 &  85 & 0.052 & 0.031 & 0.039 & 0.218 & 0.129 & 0.162 \\
Human B (longer)   & GPT-4o     & 154 & 258 & 0.066 & 0.109 & 0.082 & 0.211 & 0.349 & 0.263 \\
Human B (shorter)  & Human A (shorter)  & 161 & 144 & 0.104 & 0.107 & 0.106 & 0.280 & 0.288 & 0.284 \\
Human B (shorter)  & Gemma3             & 161 &  93 & 0.107 & 0.085 & 0.095 & 0.253 & 0.202 & 0.225 \\
Human B (shorter)  & GPT-4o   & 161 & 298 & 0.076 & 0.167 & 0.104 & 0.210 & \textbf{0.464} & 0.289 \\
\bottomrule
\end{tabular}
\caption{Evaluation metrics for reference and hypothesis campaign comparisons. Ref and Hyp Ct refer to the count of error annotations made by the respective annotation campaigns. Hard (H) metrics require span overlap and error type equivalence while soft (S) metrics only require span overlap. Maximum values are highlighted in bold.}
\label{tab:campaign_metrics_max}
\end{table*}

\vspace{-0.5em}
\paragraph{LLM-as-a-Judge}

\begin{figure}
    \centering
    \includegraphics[width=\linewidth]{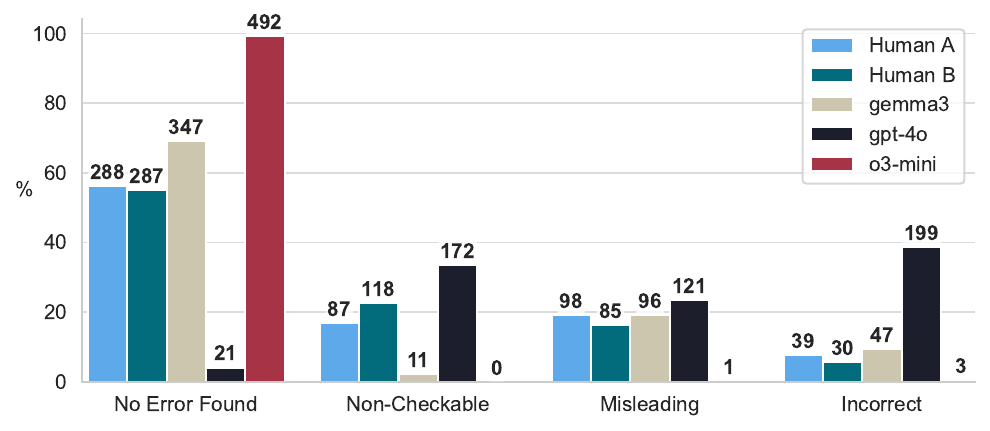}
    \caption{Distribution of error types based on human and LLM-based span annotation campaigns.}
    \label{fig:span_error_distribution}
\end{figure}

Many papers rely on LLMs for annotation, often replacing human annotation to save money.
Following \citet{kocmi-etal-2024-error} and \citet{kasner2025largelanguagemodelsspan}, we used LLMs as span annotators to identify errors in the text and to provide a reasoning for each error.

We used GPT-4o \citep{openai2024gpt4ocard}, o3-mini,\footnote{\url{https://platform.openai.com/docs/models/o3-mini}} and Gemma3 \citep{gemmateam2025gemma3technicalreport}.
As shown in Figure \ref{fig:span_error_distribution}, GPT-4o heavily overannotates, frequently citing a nonsensical reason.
This is especially notable in the \emph{incorrect} error category.
Conversely, o3-mini severely under-annotates, although the few annotations it produces are accurate.
Gemma3 most closely matches the human distribution of errors, but it under-utilizes the \emph{non-checkable} category.

We used precision, recall, and F1 to measure agreement on the example level, shown in Table~\ref{tab:campaign_metrics_max}.
The precision and F1 between the two human annotator groups set a reasonable baseline and we can see that none of the LLMs reaches this baseline yet.
On the other hand, GPT-4o achieves a high recall due to its overannotation.

Curiously, prompting the models to be more lenient and explicitly showing an example of a correct highlight led to a higher count of annotated errors.
In sum, we are able get more granular insights from human annotation, but LLMs are not yet to be fully trusted to evaluate this task.

\vspace{-0.5em}
\section{Discussion}
\vspace{-0.5em}
\paragraph{Gold Human Annotation?}
After paying a significant sum of almost £800 for human annotations and observing a poor quality of the resulting annotations, despite taking multiple precautions (pre-filtering of participants, multiple pilots, payment bonus for passing an attention check), we conclude that tasks with longer inputs that require potentially subjective judgments are not suitable to be evaluated through crowdworkers.

\vspace{-0.5em}
\paragraph{Some LLMs pass attention checks}
Gemma3 and o3-mini both passed the attention check that was designed to be simple and visible, yet only 24\% of crowdworkers actually passed it. 
This demonstrates that LLMs are capable of capturing the more visible errors, but still have room for improvement in subtler errors.

Running Gemma3 without validation would underestimate the amount of non-checkable errors which have a higher business impact.
Similarly, relying on GPT-4o without the context of validation would significantly overestimate the seriousness of incorrect errors.

\vspace{-0.5em}
\paragraph{Simpler methods to the rescue}
We observed that objective and quick to compute metrics, such as word overlap, correlate well with human judgment. 
We argue that they are a solid choice to be measured and reported in the absence of other evaluation metrics for estimating faithfulness.

\vspace{-0.5em}
\section{Conclusion}
\vspace{-0.5em}
Addressing our research questions, we demonstrated that: (1) NLI entailment and simple statistical metrics achieve moderate correlation with human judgments and are thus the best out-of-the-box options for measuring faithfulness; (2) referenceless evaluation can be effective when validated properly; (3) while LLMs excel at generating hotel highlights, they prove unreliable as evaluators of content faithfulness;
and (4) non-checkable and incorrect information have the highest potential for negative business impact.

We believe that real-world evaluation of tasks that emerged with the rise of LLMs and few-shot prompting should be studied carefully. Current evaluation methods are insufficient for automated quality assurance, and the errors that go unnoticed are likely to cause a negative business impact.

\section{Limitations}
Despite our precautions, the span annotations from crowdworkers were of a less-than-satisfactory quality.
We still used this data for the validation of LLM-as-a-judge, because it was not feasible for us to annotate sufficient quantity of data in-house. 
We believe there are still signals to be learned from this noisy data.

\section{Ethical Consideration}
\paragraph{Human Annotations}

The payment structure included an £8 per hour base rate paid out to all annotators after finishing the task -- regardless of their annotation quality.
We paid out a bonus of £4.60 per hour to workers who passed our attention check.
This ensured compliant workers received the UK living wage of £12.60 per hour.\footnote{\url{https://www.livingwage.org.uk/}}
For comparison, the Prolific minimum wage is £6.00 per hour and the recommended wage is £9.00 per hour.\footnote{\url{https://researcher-help.prolific.com/en/article/2273bd}}

\paragraph{Model Inference}
The total cost to run the LLMs for span annotations (2-3 runs on 500 examples per model to optimize the prompt) through APIs was less than \$100.

\paragraph{Use of AI}
We used AI-assisted coding (i.e. Copilot) with the bulk being human-written. For writing, AI was used to check grammar mistakes.

\bibliography{anthology,custom}

\newpage
\appendix

\section{Full Spearman Rank Correlation Results}
\label{sec:appendix}

In Table \ref{tab:hallucination_and_error_correlations}, we present Spearman rank correlations of all automatic methods with categorical human judgment (column Hallucination \%) as well as the presence of a certain error category in the span-annotated data.
In the span-annotated data, the correlations are lower, because they are compared to discrete values:
\begin{compactitem}
    \item 0 of the two annotators found the given error type
    \item 1 of the annotators found the given error type
    \item 2 -- both annotators found the given error type
\end{compactitem}

\begin{table*}[htbp]
\centering
\small
\begin{tabular}{llrrrrr}
\toprule
\textbf{Type} & 
\textbf{Metric} & \textbf{Hallucination \%} & \textbf{Non-Checkable} & \textbf{Misleading} & \textbf{Incorrect} & \textbf{Any Error} \\
\midrule
Trainable & BERTScore F1          & -0.39 & -0.11 &  0.06 & -0.04 & -0.06 \\
Trainable & BERTScore Precision   & -0.58 & -0.13 &  0.06 & -0.02 & -0.08 \\
Trainable & BERTScore Recall      & -0.24 & -0.09 &  0.05 & -0.05 & -0.04 \\
Trainable & MBERT Contradiction   &  0.36 &  0.11 &  0.06 &  0.05 &  0.15 \\
Trainable & MBERT Neutral         &  0.66 &  \textbf{0.24} &  0.05 &  0.04 &  \textbf{0.22} \\
Trainable & MBERT Entailment      & \textbf{-0.67} & -0.23 & -0.06 & -0.04 & \textbf{-0.22} \\
Trainable &  Cosine Similarity  &  0.00 & -0.02 &  \textbf{0.11} & -0.04 &  0.04 \\
Trainable &  Dot Score       & -0.04 & -0.04 &  0.07 & -0.03 &  0.01 \\
Trainable &  MPNet Similarity   & -0.00 &  0.04 &  0.07 &  0.03 &  0.08 \\
Trainable &  NegMPNet Sim.  & -0.01 &  0.05 &  0.01 & -0.05 &  0.01 \\
Trainable &  LaBSE Similarity  & -0.12 &  0.02 &  0.06 &  0.02 &  0.05 \\

\hline
Word Overlap & Noun Coverage     & -0.41 & -0.13 & -0.05 &  0.02 & -0.10 \\
Word Overlap & Adjective Coverage     & -0.55 & \textbf{-0.17} & \textbf{-0.09} & -0.01 & \textbf{-0.18} \\
Word Overlap & Verb Coverage     & -0.18 & -0.01 & -0.04 & -0.05 & -0.08 \\
Word Overlap & Form Coverage     & \textbf{-0.63} & -0.16 & \textbf{-0.09} & -0.05 & \textbf{-0.17} \\
Word Overlap & Lemma Coverage    & -0.62 & -0.16 & -0.07 & -0.01 & -0.15 \\
Word Overlap & Entity Coverage   & -0.05 &  0.01 &  0.04 & \textbf{-0.10} & -0.02 \\
Other & Num. of Entities  &  0.01 &  0.01 &  0.08 &  0.09 &  0.09 \\
\hline
N-gram Overlap & BLEU       & -0.51 & -0.08 & -0.03 & -0.04 & -0.09 \\
N-gram Overlap & ROUGE-1 P  & -0.34 & -0.07 &  0.06 & -0.04 & -0.05 \\
N-gram Overlap & ROUGE-1 R  & \textbf{-0.58} & \textbf{-0.16} & \textbf{-0.09} & -0.04 & \textbf{-0.18} \\
N-gram Overlap & ROUGE-1 F  & -0.37 & -0.07 &  0.06 & -0.04 & -0.05 \\
N-gram Overlap & ROUGE-L P  & -0.39 & -0.10 &  0.06 & -0.04 & -0.06 \\
N-gram Overlap & ROUGE-L R  & \textbf{-0.56} & \textbf{-0.18} & -0.06 & -0.02 & \textbf{-0.15} \\
N-gram Overlap & ROUGE-L F  & -0.41 & -0.11 &  0.06 & -0.04 & -0.06 \\
\bottomrule
\end{tabular}
\caption{Correlations between various metrics and hallucination percentage obtained from categorical human data. Also shown are correlations with span-annotated human data: Non-Checkable, Misleading, Incorrect, and Any Error. }
\label{tab:hallucination_and_error_correlations}
\end{table*}

\section{Implementation Details}
We used the sacrebleu package to compute the frequencies of 1- to 4-grams. We then computed the harmonic mean ourselves to avoid the brevity penalty, that was undesirable in our case.
To compute Rouge scores, we used the rouge\_score Python package.
For trainable metrics, we used models available on HuggingFace:
\begin{itemize}
    \item msmarco-distilbert-base-v4 \citep{reimers-gurevych-2019-sentence}\footnote{\url{https://huggingface.co/sentence-transformers/msmarco-distilbert-base-v4}}
    \item msmarco-distilbert-base-tas-b \citep{reimers-gurevych-2019-sentence}\footnote{\url{https://huggingface.co/sentence-transformers/msmarco-distilbert-base-tas-b}}
    \item sentence-transformers/all-mpnet-base-v2 \citep{reimers-gurevych-2019-sentence}\footnote{\url{https://huggingface.co/sentence-transformers/all-mpnet-base-v2}}
    \item tum-nlp/NegMPNet \citep{anschutz-etal-2023-correct}\footnote{\url{https://huggingface.co/tum-nlp/NegMPNet}}
    \item sentence-transformers/LaBSE \citep{feng-etal-2022-language} \footnote{\url{https://huggingface.co/sentence-transformers/LaBSE}}
    \item tasksource/ModernBERT-base-nli \citep{sileo-2024-tasksource}\footnote{\url{https://huggingface.co/tasksource/ModernBERT-base-nli}}
    \item cross-encoder/nli-deberta-v3-base \citep{he2021deberta} \footnote{\url{https://huggingface.co/cross-encoder/nli-deberta-v3-base}}
\end{itemize}
We use the implementation of BertScore from HuggingFace's evaluate library.
For part-of-speech tagging and named entity recognition, we used SpaCy \citep{honnibal2020spacy}.

\section{Annotator Instructions}
We present the instructions given to annotators in Table \ref{tab:instructions}.

\begin{table}[htbp]
\centering
\caption{Definitions and Examples of Error Types}
\label{tab:instructions}
\begin{tabular}{|p{0.15\textwidth}|p{0.35\textwidth}|p{0.4\textwidth}|}
\hline
\textbf{Error Type} & \textbf{Definition} & \textbf{Example} \\
\hline
Not Checkable & The summary contains information that is not mentioned anywhere in the original text. This information could either be objective (such as the presence of a swimming pool) or subjective (such as quietness). & 
\textbf{Text:} \textit{A fun-filled vacation or relaxing business trip awaits you at the Holiday Inn Express \& Suites Tampa Airport nestled on the beautiful waters of Tampa Bay at Rocky Point. Our hotel is minutes from the beautiful waterfront views of Tampa's Famous Riverwalk featuring miles of shops, artists and Tampa's premier dining. Our friendly and knowledgeable staff invite you to relax in the outdoor pool.}

\textbf{Summary:} Enjoy stunning views of Tampa Bay and the beautiful waterfront from this \underline{pet-friendly} hotel.

\textbf{Explanation:} It was not mentioned whether the hotel is pet-friendly, thus this information is Not Checkable. \\
\hline
Misleading & The summary presents information that appears in the original text, however, it does so in a way that changes the perceived meaning. This can be due to subjective judgments (is an attraction 10 km away ``close''?) or due to a word that can have multiple meanings (pool as in swimming pool or the game requiring a pool table). & 
\textbf{Text:} \textit{Sheraton Düsseldorf Airport hotel is directly connected with the Terminal - in the unique location on the roof of car park P3, surrounded by 10,000m² greenery. [...] Relax from your travels or prepare for your meeting with green views.}

\textbf{Summary:} Enjoy breathtaking views from the rooftop \underline{terrace and garden}, offering a relaxing escape.

\textbf{Explanation:} Terrace and garden are Misleading. The hotel seems to be on the roof, but there is no mention of a terrace. At the same time, 10,000m² seems unlikely to be a garden. \\
\hline
Incorrect & The summary contains information that either contradicts a statement from the original text (i.e the text mentioning the hotel is NOT pet-friendly, but the summary stating it is) or contains a severe error, such as using a wrong entity (e.g. place or a person), or a wrong number (for example confusion of different numbers or kilometers vs miles). & 
\textbf{Text:} \textit{The closest major airports to Bomontist Suit are: Istanbul (SAW-Sabiha Gokcen Intl.) - 17.5 km / 10.9 mi Istanbul (IST-Ataturk Intl.).}

\textbf{Summary:} \underline{Schiphol Airport} is just a \underline{15-minute drive} from the hotel.

\textbf{Explanation:} Schiphol Airport in Amsterdam is Incorrect, since the accommodation is clearly in Istanbul. In addition, 15-minute drive is Not Checkable in this context, because even though we know the distance, we don't know the expected speed of the journey. \\
\hline
\end{tabular}
\end{table}

\end{document}